\documentclass{article}
\pdfoutput=1
    \PassOptionsToPackage{numbers, compress, sort, comma, square}{natbib}

    \usepackage[preprint]{neurips_2020}

\usepackage[utf8]{inputenc} 
\usepackage[T1]{fontenc}    
\usepackage{hyperref}       
\usepackage{url}            
\usepackage{booktabs}       
\usepackage{amsfonts}       
\usepackage{nicefrac}       
\usepackage{microtype}      
\usepackage{amsmath}
\usepackage[ruled,vlined]{algorithm2e}
\usepackage{graphicx}
\usepackage{float}
\usepackage{wrapfig}

\title{Amended Cross Entropy Cost: Framework For Explicit Diversity Encouragement}

\author{%
    Ron Shoham and Haim Permuter\\
    Department of Electrical and Computer Engineering\\
    Ben Gurion University\\
    \texttt{ronshoh@post.bgu.ac.il, haimp@bgu.ac.il} \\
}

\begin{document}

\maketitle
\begin{abstract}
Cross Entropy (CE) has an important role in machine learning and, in particular, in neural networks. It is commonly used in neural networks as the cost between the known distribution of the label and the Softmax/Sigmoid output. In this paper we present a new cost function called the Amended Cross Entropy (ACE). Its novelty lies in its affording the capability to train multiple classifiers while explicitly controlling the diversity between them. We derived the new cost by mathematical analysis and  ``reverse engineering'' of the way we wish the gradients to behave, and produced a tailor-made, elegant and intuitive cost function to achieve the desired result. This process is similar to the way that CE cost is picked as a cost function for the Softmax/Sigmoid classifiers for obtaining  linear derivatives. By choosing the optimal diversity factor we produce an ensemble which yields better results than the vanilla one. We demonstrate two potential usages of this outcome, and present empirical results. Our method works for classification problems analogously to Negative Correlation Learning (NCL) for regression problems.
\end{abstract}
\section{Introduction and motivation}
It has been shown in several studies, both theoretically and empirically, that training an ensemble of models, i.e. aggregating predictions from multiple models, is superior to training a single model\cite{brown2005managing,NIPS2018_7614,Fernandez2014hundreds,NIPS2019_9097,liu1999ensemble,ren2016ensemble, NIPS1995_1044, NIPS2019_8443, Zhang2017ensemble, adaboost}. Many works point out that one of the keys for an ensemble to perform well is to encourage diversity among the models \cite{acc_and_div_ens,NIPS2016_6289,NIPS2016_6270,liu1999ensemble,shi2018crowd,NIPS2018_7831,adaboost}. This property is the main motivation our work. 
\par Sigmoid and Softmax are both well known functions which are used for classification (the former for binary and the second for multi label classifications). Both are used to generate distribution vectors $q_Y(x)=\{q_1(x),..,q_L(x)\}$ over the labels $Y=\{1,..,L\}$, where $x$ is a given input. For Deep Neural Networks (DNNs) the framework of applying a Sigmoid/Softmax on top of the network is very popular, where the goal is to estimate the real distribution $p_Y(x)=\{p_1(x),..,p_L(x)\}$, which might be a 1-hot vector for a hard label. Henceforth, we omit $x$ unless it is crucial for some definition or proof. We denote $p=p_Y(x),q=q_Y(x)$. We optimize $q$ by minimizing the CE cost function 
\begin{align}
    H(p,q)&=E_p[-\log q] \nonumber\\
    &=-\sum_{i=1}^L p_i\log q_i.\label{CE}
\end{align} 
The optimization is usually gradient based\cite{kingma2014adam, RMSProp}. Hence, one of the main motivations for using the CE cost function over Sigmoid/Softmax outputs is the linear structure of the gradient, which is similar to that obtained by applying the Mean Squared Error (MSE) method over a linear regression estimator. Studies show that this property is important for preventing vanishing gradient phenomena \cite{Goodfellow-et-al-2016,nielsen2015neural}.
\par Now let us define the setting of the ensemble problem. We train $K$ classifiers, with distribution functions $q^1,..,q^K$, to generate ensemble $\overline{q}=\frac{1}{K}\sum_{k=1}^Kq^k$, which estimates the real distribution $p$. This setting is very common and the straightforward way to tackle it is by training each model independently using the CE cost function $H(p,q^k)$. Encouraging diversity is manifested by using different training samples or different seeds for weight initialization. However, to the best of our knowledge, there is no explicit way to control the  ``amount'' of diversity between the classifiers. 
\par In this work we present a novel framework, called Amended Cross Entropy (ACE), which makes it possible for us to train each model and, simultaneously, to achieve diversity between the classifiers. Our main result in this work is the introduction of a new cost function
\begin{align}
     H(p,q^k)-\frac{\gamma}{K-1}\sum_{j\neq k}H(q^j,q^k),
\end{align}
which is applied for the $k$-th classifier and is not independent of the other classifiers. We see that ACE is built from the vanilla CE between $p$ and $q^k$, minus the average of the CE between $q^k$ and the other estimators, factored with $\gamma$. This result is very intuitive since we wish to minimize the CE of the estimated distribution with the real one, while enlarging the CE of the estimator with the others, i.e. encourage diversity. The hyper-parameter $\gamma\in [0,\frac{K-1}{K}]$ explicitly controls the diversity, and is fine-tuned in order to achieve optimal results. The development of ACE starts from an assumption of the structure we wish the gradient to be in. As we show in this paper, a similar assumption lies at the base of applying CE over Softmax. We develop a variant especially for DNNs, which can be stacked on top of the network instead of the vanilla Softmax layer, and makes it possible to yield superior results without significantly increasing the number of parameters or the computational resources.
\par This work has been inspired by the Negative Correlation Learning (NCL) \cite{brown2005managing,liu1999ensemble,shi2018crowd} framework, which is used for regression ensembles. In the next section we will present the NCL framework, its development and its results, in order to explain the analogous approach we used in our work.
\section{Related work: Negative Correlation Learning (NCL)}
\citet{liu1999ensemble} and \citet{brown2005managing} presented the NCL framework as a solution for the diversity issue for ensembles of regression. Let us denote $X$ as the vector of features and $Y$ as the target. The goal is to find $F:\mathcal{X}\rightarrow \mathcal{Y}$ which yields as low as possible error w.r.t. MSE criteria, i.e. to minimize 
\begin{align}
    e(F) = \int (F(X, \theta)-Y)^2p(X,Y)d(X,Y).\label{expected_square_error}
\end{align}
Here, $\theta$ stands for the parameters of $F$. In practice, the distribution $p(X,Y)$ is unknown, so we use $N$ realizations (training set) $\{(x_1,y_1),..(x_N,y_N)\}$ to estimate \eqref{expected_square_error} with an empirical MSE using $\hat{e}(F)=\frac{1}{N}\sum^N_{i=1}(F(x_i, \theta)-y_i)^2$.
Under the assumption that $(X_i,Y_i)$ are i.i.d., or at least stationary and ergodic, $\hat{e}(F)$ converges to $e(F)$.
We use the short notation $F$ to denote $F(X,\theta)$. Instead of \eqref{expected_square_error} we can use the expectation operator $E$ and decompose the error to the known structure of bias and variance
\begin{align}
    E[(F-Y)^2] &= (E[F]-Y)^2 + E[(F-E[F])^2]\nonumber\\
               &= \textit{bias(F)}^2 + \textit{variance(F)}.
\end{align}
A common way to apply an ensemble of models is to average multiple trained estimators $\{F^1,..,F^K\}$ 
\begin{align}
    \overline{F} = \frac{1}{K}\sum^K_{k=1}F^k.
\end{align}
By checking the decomposition of the ensemble expected error it is straightforward to show that 
\begin{align}
    E[(\overline{F}-Y)^2] = &(E[\overline{F}]-Y)^2 + E[(\overline{F}-E[\overline{F}])^2]\nonumber\\
    = &\frac{1}{K^2}\sum^K_{k=1}(E[F^k] - Y)^2 + \frac{1}{K^2}\sum^K_{k=1}E[(F^k-E[F^k])^2]\nonumber \\
    &+ \frac{1}{K^2}\sum^K_{k=1}\sum_{j\neq k}E[(F^k-E[F^k])(F^j-E[F^j])]\nonumber\\
    = &\overline{\textit{bias}}(F)^2 + \overline{\textit{variance}}(F) + \overline{\textit{covariance}}(F).
\end{align}
This outcome is called the \textit{bias-variance-covariance} decomposition, and is the main motivation for NCL. We notice that by reducing the correlation between the estimators of an ensemble, the ensemble might yield a lower error. Based on this, \citet{liu1999ensemble} proposed a regularization factor that is added to the cost function of any of the single estimators during the training phase. This factor is an estimation of the sum of covariances between the trained estimator and the others. The factor is multiplied by a hyper-parameter $\gamma$, which explicitly controls the ``amount'' of the diversity between the single estimator and the other estimators in the ensemble
\begin{align}
    e^k &= \frac{1}{2}(F^k-Y)^2 + \gamma (F^k-\overline{F})(\sum_{j\neq k}(F^j-\overline{F}))\nonumber\\
    &= \frac{1}{2}(F^k-Y)^2 - \gamma(F^k-\overline{F})^2.\label{distance1}
\end{align}
Notice that in order to avoid a factor of $2$ in the gradient analysis, we multiply the MSE by a factor of $\frac{1}{2}$. By setting $\gamma=0$ we get the conventional MSE cost function, i.e. each model is optimized independently.
\paragraph{Gradient analysis} 
Gradient-wise optimization\cite{kingma2014adam, RMSProp} is a very popular method for optimizing a model. Therefore, conducting analysis over the gradient behaviour of a cost function is advisable. Let us check the gradient of the cost function $e^k$ with respect to $F^k$
\begin{align}
    \frac{\partial e^k}{\partial F^k} &= (F^k-Y) - \gamma[2(1-\frac{1}{K})(F^k-\overline{F})].
\end{align}
By defining $\lambda=2\gamma(1-\frac{1}{K})$, we get
\begin{align}
    \frac{\partial e^k}{\partial F^k}&=(F^k-Y) - \lambda(F^k-\overline{F}) \nonumber\\
    &= (1-\lambda)(F^k-Y) + \lambda(\overline{F}-Y) \label{NCL_gradient}.
\end{align}
We notice again that by setting $\gamma=\lambda=0$ we get the same gradient as with independent training.
\subsection{Usage of NCL}
\begin{figure}[!ht]
  \centering
   \includegraphics[clip, trim=3.2cm 12.5cm 0.5cm 9.8cm,width=0.9\linewidth]{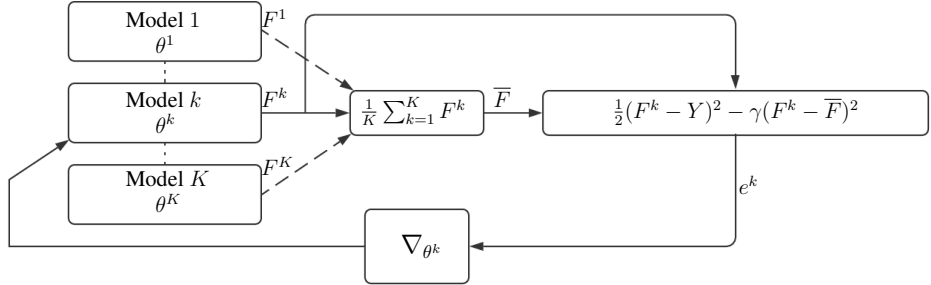}
  \caption{NCL. A sketch of a training phase of the $k$-th model. First, the input is processed by $K$ models, which yields the predictions $\{F^1,..,F^K\}$. Using this, the cost function $e^k$ is calculated. Finally, the gradient of $\theta^k$ is calculated and model $k$ is updated accordingly.}
  \label{Regular}
\end{figure}

\citet{liu1999ensemble} and \citet{brown2005managing} suggested a vanilla approach for optimizing multiple regressors. They suggested training multiple regression models that do not have to be of the same architecture, but train simultaneously in order to reduce the correlation between the models. The architecture is presented in Fig.~\ref{Regular}.
However, applying this approach, the computational power and the number of parameters used increases significantly. For example, if we use the same architecture for all of the $K$ models, we use $K$ times the number of parameters used by a single model. If we train a DNN with millions of parameters, this might result in a non scalable training scheme.
\par In order to handle this, \citet{shi2018crowd} suggested a new approach. They suggested stacking a layer of a regressors ensemble on top of a DNN instead of the vanilla regression layer. In this way, they claimed that they got the benefit of NCL while not increasing the number of parameters and computational power significantly. This architecture, called \textit{D-ConvNet}, yields state of the art results in a \textit{Crowd Counting} task. The work, as well as a sketch of the architecture can be seen in their paper \cite{shi2018crowd}.
\section{Amended Cross Entropy (ACE)}
In this section we first show the main motivation for using the CE cost function for a Softmax classifier. Like many other functions, CE achieves its minima when both of the distribution vectors are equal (MSE, Mean Absolute Error (MAE), etc.). However, CE is the only cost function which yields a linear gradient for a distribution generated by Softmax, similarly to the gradient of the MSE cost function over a linear regressor. We show this over a single classifier case first, and later we use this approach analogously for multi-classifiers, where we wish to yield the same gradient structure as in NCL, in order to analytically develop the ACE framework for multi-classifiers. 
\paragraph{CE cost function for Softmax classifier} 
Let us denote $L$ as the size of the set of events (labels), and $p=\{p_1,..,p_L\}$ as the real distribution vector for a given input (which is a 1-hot vector for a hard label). We wish to train an estimator $q=\{q_1,..,q_L\}$ for the real distribution. We denote the estimator parameters as $\theta$. The estimator generates a raw vector $z=\{z_1,..,z_L\}$, which is a function of the input, and applies Softmax $\sigma(z)$ over it in order to yield the estimator $q$, i.e.
\begin{align}
    q &= \sigma(z) \nonumber\\
    &= \left\{\frac{e^{z_1}}{\sum_{l=1}^Le^{z_l}}, \dots ,\frac{e^{z_L}}{\sum_{l=1}^Le^{z_l}}\right\} \nonumber\\
    &=\{q_1,..,q_L\}.
\end{align}
Later, a CE cost function is applied to measure the error between the estimator and the real distribution \eqref{CE}. In order to optimize the estimator's parameters $\theta$, gradient based methods are applied\cite{kingma2014adam, RMSProp}. The gradient is calculated using the chain rule
\begin{align}
    \nabla_\theta H(p,q) &= \nabla_\theta z \nabla_z H(p,q).
\end{align}
Now, let us calculate $\nabla_z H(p,q)$ explicitly
\begin{align}
    \nabla_z H(p,q) &= \nabla_z \left( -\sum_{i=1}^L p_i\log q_i\right) \nonumber\\
    &= \nabla_z\left( -\sum_{i=1}^L p_i\log \frac{e^{z_i}}{\sum_{l=1}^Le^{z_l}}\right) \nonumber\\
    &= \left\{ \frac{\partial}{\partial z_1}\left(-\sum_{i=1}^L p_i\log \frac{e^{z_i}}{\sum_{l=1}^Le^{z_l}}\right),..,\frac{\partial}{\partial z_L}\left(-\sum_{i=1}^L p_i\log \frac{e^{z_i}}{\sum_{l=1}^Le^{z_l}}\right)   \right\}\nonumber\\
    &= \left\{ \frac{e^{z_1}}{\sum_{l=1}^Le^{z_l}} - p_1, \dots ,\frac{e^{z_L}}{\sum_{l=1}^Le^{z_l}}-p_L      \right\} \nonumber \\
    &= \left\{q_1 - p_1, \dots, q_L-p_L  \right\} \nonumber\\
    &= q-p.\label{gradient1}
\end{align}
We see that a linear structure of a gradient is obtained when applying CE over a Softmax classifier. This structure is similar to that of the MSE cost function over a linear regression estimator\cite{ Goodfellow-et-al-2016,nielsen2015neural}.
\subsection{ACE}
Inspired by the NCL result and by our belief that an important consideration for the choice of a cost function is the gradient behaviour (as long as it is a valid cost function), we wish to find a cost function that would yield the same properties. Therefore, we first assume the gradient structure, and later integrate it in order to find the appropriate cost function. Let us denote $K$ as the number of classifiers in the ensemble, $e^k$ as the $k$-th model cost function, $z^k$ as the raw output vector of the $k$-th model, $q^k=\sigma(z^k)$ as the estimated distribution of the $k$-th model, and $\theta^k$ as the parameters of the $k$-th model. We would like to train an ensemble of models $\overline{q} = \frac{1}{K}\sum_{k=1}^K q^k$ to estimate $p$. Since the gradient structure might be one of the most important considerations for choosing and constructing a cost function, by combining the results of \eqref{NCL_gradient} and \eqref{gradient1} we assume a gradient 
\begin{align}
    \nabla_{z^k}\, e^k &= (1-\lambda)(q^k-p) + \lambda (\overline{q}-p)\nonumber\\
    &= (q^k-p) -\frac{\lambda}{K}\sum_{j\neq k}(q^k-q^j).\label{gradient2}
\end{align}
This assumption is the foundation of our proposed method and is the basis for developing the ACE framework.
In order to find $e^k$ we need to integrate the above with respect to $z^k$
\begin{align}
    e^k &= \int \left( (q^k-p) -\frac{\lambda}{K}\sum_{j\neq k}(q^k-q^j) \right) dz^k \nonumber\\
    &=\int \left(q^k-p   \right) dz^k -\frac{\lambda}{K}\sum_{j\neq k}\int \left(q^k-q^j   \right) dz^k.\label{integral1}
\end{align}
By reverse engineering \eqref{gradient1}, and using the fact that $p$ and $q^j,\; \forall j\neq k$ are independent of $z^k$, we get
\begin{align}
    e^k &= H(p,q^k)-\frac{\lambda}{K}\sum_{j\neq k}H(q^j,q^k) + C, \label{result_amended}
\end{align}
where $C$ is a constant independent of $z^k$. We set $C=0$. We can also set $\gamma=\lambda\frac{K-1}{K}$ in order to get
\begin{align}
         H(p,q^k)-\frac{\gamma}{K-1}\sum_{j\neq k}H(q^j,q^k),
\end{align}
i.e. the average of the CE between the $k$-th classifier and the others. Notice that by setting $\lambda=\gamma=0$ we get the regular CE cost function.
\paragraph{Alternative formulation and analogy to NCL} 
Using algebraic manipulations, one can show that ACE \eqref{result_amended} has a similar structure to the one of NCL \eqref{distance1}. Let us check the result in \eqref{result_amended}
\begin{align}
    e^k &= H(p,q^k)-\frac{\lambda}{K}\sum_{j\neq k}H(q^j,q^k)\nonumber\\
    &= H(p,q^k) - \lambda H(\overline{q},q^k) +\frac{\lambda}{K} H(q^k,q^k).\label{nicer_ACE}
\end{align}
Note that $H(q^k,q^k)=H(q^k)$, i.e. the entropy of $q^k$. Now let us check the result in \eqref{distance1}
\begin{align}
    e^k &= \frac{1}{2}(F^k-Y)^2 - \gamma(F^k-\overline{F})^2\nonumber\\
    & = \frac{1}{2}(F^k-Y)^2 - \gamma(F^k-\overline{F})^2 + (F^k-F^k)^2.
\end{align}
If we refer to the MSE and CE as divergence operators $D_{MSE}$ and $D_{CE}$, respectively, we can observe that both of the cost functions have the same structure
\begin{align}
    e_{NCL}^k &= a_1 D_{MSE}(F^k,Y) - a_2 D_{MSE}(F^k,\overline{F}) + a_3 D_{MSE}(F^k,F^k),\label{distance3}\\
   e_{ACE}^k &= b_1 D_{CE}(q^k,p) - b_2 D_{CE}(q^k,\overline{q}) + b_3 D_{CE}(q^k,q^k),\label{distance4}
\end{align}
where $a_i,b_i$ are constants. The first component of both expressions in \eqref{distance3} and \eqref{distance4} is the divergence between the real value and the estimator's prediction, i.e. the vanilla error. The second component is a negative divergence between estimator's prediction and the ensemble prediction. Minimizing it (maximizing the divergence) encourages diversity between the estimator and the ensemble. The last component is the minimum of the divergence, where for MSE it is zero and for CE it is the entropy.  
\paragraph{Non-uniform weights} Let us check the case where our ensemble is aggregated using non-uniform weights, i.e. $\overline{q} = \sum_{k=1}^K \alpha^k q^k$, where $\alpha^k\geq 0$, $\forall k$, and $\sum_{k=1}^K \alpha^k=1$. Instead of \eqref{gradient2} we get 
\begin{align}
    \nabla_{z^k}\, e^k &= (1-\lambda)(q^k-p) + \lambda (\overline{q}-p)\nonumber\\
    &= (q^k-p) -\lambda\sum_{j\neq k}\alpha^j(q^k-q^j).\label{gradient3}
\end{align}
Hence, for weights $\alpha^1,..,\alpha^K$ which are independent of $z^k$, instead of \eqref{result_amended} we obtain
\begin{align}
    e^k &= H(p,q^k)-\lambda\sum_{j\neq k}\alpha^j H(q^j,q^k).\label{non_uniform_ACE}
\end{align}
\section{Implementation}
In this section we examine two alternative implementations for the result we got above.
\subsection{ACE for multiple models}
\begin{figure}[!ht]
  \centering
  \includegraphics[clip, trim=3.2cm 12.5cm 0.5cm 9.8cm,width=0.9\linewidth]{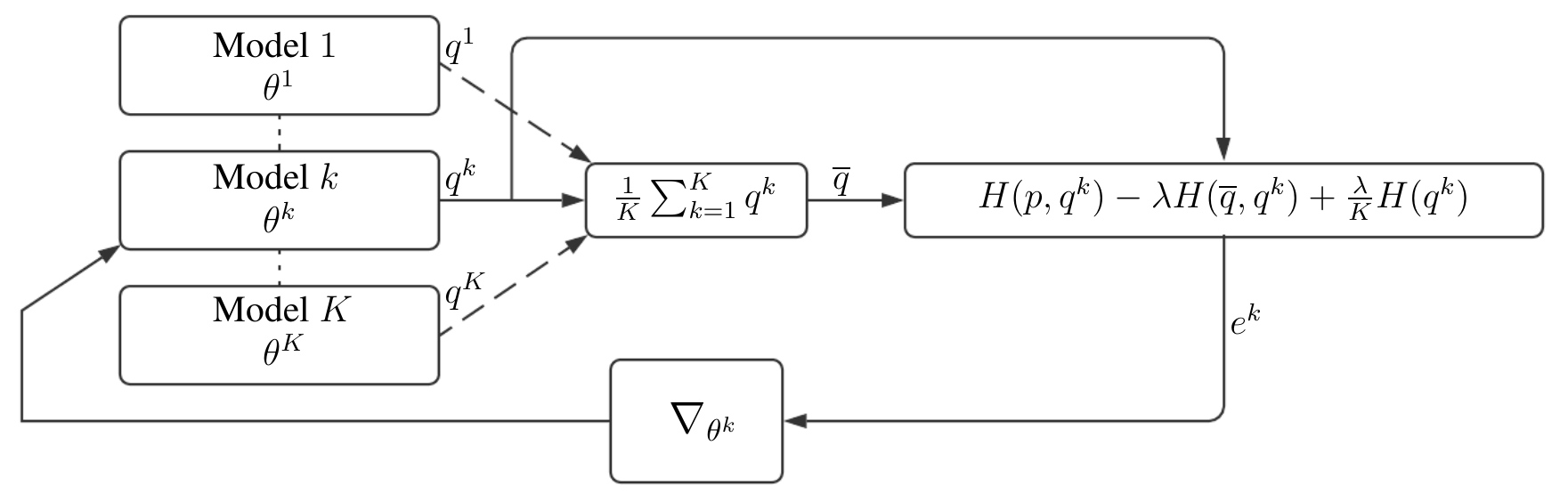}
  \caption{ACE for multiple models. A sketch of a training phase of the $k$-th model. First, the input is processed by $K$ models, which yields the distribution vectors $\{q^1,..,q^K\}$. Later, the cost function $e^k$ is calculated. Finally, the gradient of $\theta^k$ is calculated and model $k$ is updated accordingly.}
  \label{vanilla}
\end{figure}

The straightforward vanilla implementation of our result is training multiple models simultaneously using ACE. In this approach we train $K$ models and fine-tune $\lambda$ to yield the best ensemble result.
  \begin{wrapfigure}{r}{0.57\textwidth}
    \begin{minipage}{0.57\textwidth}
      \begin{algorithm}[H]
            \SetAlgoLined
             \For{$k$ in $\{1,..,K\}$}{
              calculate predictions $q^k$\;
             }
             \For{$k$ in $\{1,..,K\}$}{
              calculate loss $e^k$ \eqref{nicer_ACE}\;
              calculate gradient $\nabla_{\theta^k}e^k$\;
              apply optimization step over $\theta^k$ using $\nabla_{\theta_k}e^k$\;
             }
             \label{algo_1}
              \caption{Training step of ACE for $K$ models with respect to a single input with probability vector $p$}
         \end{algorithm}
    \end{minipage}
  \end{wrapfigure}
The models do not have to be of the same architecture. Let us denote $\theta^1,..\theta^K$ as the parameters of the models $q^1,..,q^K$, respectively. The loss functions $e^k$ are calculated as in \eqref{nicer_ACE}. We calculate the gradient for each parameter set $\theta^k$ with respect to the corresponding loss function $e^k$ (Algorithm \ref{algo_1}). This can also be used over a batch of samples while averaging the gradients. A sketch of this architecture can be viewed in Fig.~\ref{vanilla}. In the inference phase, we calculate the outputs of all of the models, and average them to yield a prediction. 
\subsection{Stacked Mixture Of Classifiers}
\begin{figure}[!ht]
  \centering
   \includegraphics[clip, trim=0.55cm 12.2cm 0.5cm 9.8cm,width=0.99\linewidth]{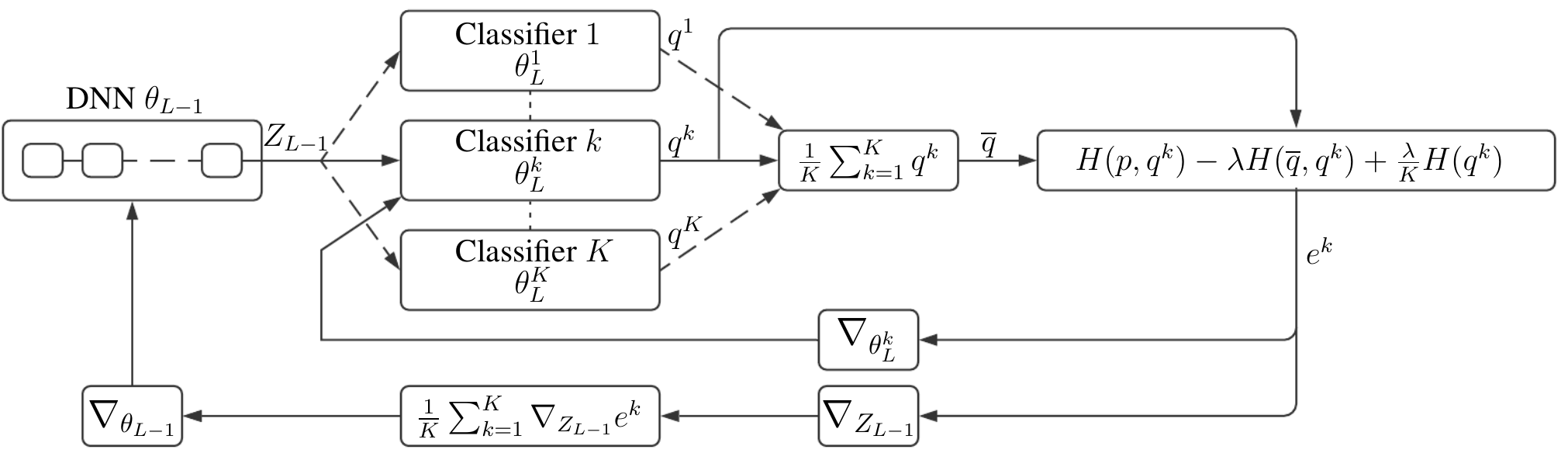}
  \caption{SMOC. A sketch of a training phase of the $k$-th classifier. First, the input is processed by a DNN, which generates $Z_{L-1}$. Second, $Z_{L-1}$ is processed by a pool of classifiers, which yields the distribution vectors $\{q^1,..,q^K\}$. Each classifier is optimized by its corresponding ACE cost function $e^k$. The gradient w.r.t. $\theta_L^k$ is calculated and classifier $k$ is updated accordingly. The gradient w.r.t. $Z_{L-1}$ is calculated and later the $K$ gradients are averaged and used to calculate the gradient w.r.t. $\theta_{L-1}$ \eqref{chain_theta}.}
  \label{smoc}
\end{figure}

A drawback of the above usage is that it takes $K$ times the computational power and memory compared to training a single vanilla model. In order to avoid this overhead and to still gain the advantages of training multiple classifiers using ACE we developed a new architecture called Stacked Mixture Of Classifiers (SMOC). This implementation is an ad-hoc variant for DNNs. Let us denote $L$ as the depth of a DNN, and $Z_{L-1}$ as the output vector of the first $L-1$ layers of the net. Usually, we stack a fully-connected layer and Softamx activation on top of $Z_{L-1}$ such that $q = \sigma(wZ_{L-1} + b)$, where $w$ and $b$ are the matrix and the bias of the last fully-connected layer, respectively, and $q$ is the output of the DNN. Instead, we stack a mixture of $K$ fully-connected+Softmax classifiers, and train them with respect to $K$ different loss functions. The output of each classifier is $q^k = \sigma(w^kZ_{L-1} + b^k)$, where $w^k$ and $b^k$ are the matrix and the bias of the $k$-th fully-connected final layer. For optimization we use ACE loss \eqref{nicer_ACE}. In the inference phase we use an average of the $K$ classifiers $\overline{q} = \frac{1}{K}\sum_{k=1}^Kq^k$. We denote this architecture as Stacked Mixture Of Classifiers (SMOC). A sketch of SMOC can be seen in Fig.~\ref{smoc}. The parameters vector $\theta_L^k$ is the set of parameters of the $k$-th final layer, i.e. $\theta_L^k=\{w^k,b^k\}$. As we can see, the number of parameters is increased by $|\theta_L^k|\times (K-1)$ compared to a similar DNN with a vanilla final layer. Using this approach, we can gain a highly diversified ensemble without having to train multiple models and increase the number of parameters significantly. Instead, we use a regular single DNN of $L-1$ layers, and create an ensemble by training multiple fully-connected+Softmax layers over its output. 
\paragraph{SMOC gradient calculation optimization}
We can think about this architecture as training $K$ DNNs which share the parameters of the first $L-1$ layers. Let us denote the shared parameters as 
  \begin{wrapfigure}{r}{0.6\textwidth}
    \begin{minipage}{0.6\textwidth}
        \begin{algorithm}[H]
          \SetAlgoLined
             calculate $Z_{L-1}$\;
             \For{$k$ in $\{1,..,K\}$}{
              calculate predictions $q^k$\;
             }
             \For{$k$ in $\{1,..,K\}$}{
              calculate loss $e^k$ \eqref{nicer_ACE}\;
              calculate gradient $\nabla_{\theta_L^k}e^k$\;
              calculate gradient $\nabla_{Z_{L-1}}e^k$\;
             }
             calculate $g(\theta_{L-1})$ \eqref{chain_theta}\;
            apply optimization step for $\{\theta_{L-1}, \theta_L^1,..,\theta_L^K\}$ using $\{g(\theta_{L-1}),\nabla_{\theta_L^1}e^1 ,.., \nabla_{\theta_L^K}e^K\}$ respectively\;
             \caption{Training step of SMOC with K stacked classifiers w.r.t. a single input with probability vector $p$}
              \label{algo_2}
         \end{algorithm}
    \end{minipage}
  \end{wrapfigure}
$\theta_{L-1}$. Similar to ACE for multiple models, we need to calculate $K$ losses and the gradients with respect to them. A naive way to do so would be to calculate the gradients separately for each cost function and to average them over the shared parameters $\theta_{L-1}$. However, this
computation has the same complexity as training $K$ different models. Since the gradients are calculated using the chain 
rule (back-propagation) we can use it to tackle this issue. Let us denote $g(\theta_{L-1})$ as the average of the gradients over the shared parameters
\begin{align}\label{average}
    g(\theta_{L-1}) = \frac{1}{K}\sum_{k=1}^{K}\nabla_{\theta_{L-1}}e^k.
\end{align}
By using the chain rule we get
\begin{align}\label{chain}
   \nabla_{\theta_{L-1}}e^k &= \nabla_{\theta_L^k}e^k\cdot\nabla_{Z_{L-1}}\theta_L^k\cdot\nabla_{\theta_{L-1}}Z_{L-1} .
\end{align}
By combining \eqref{average} and \eqref{chain}, and due to the linearity of the gradient we get
\begin{align}
   g(\theta_{L-1}) &=\frac{1}{L}\sum_{k=1}^K\nabla_{\theta_{L-1}}e^k\nonumber\\
   &= \frac{1}{K}\sum_{k=1}^K (\nabla_{\theta_L^k}e^k\cdot\nabla_{Z_{L-1}}\theta_L^k\cdot\nabla_{\theta_{L-1}}Z_{L-1})\nonumber\\
   &= \left(\frac{1}{K}\sum_{k=1}^K (\nabla_{\theta_L^k}e^k\cdot\nabla_{Z_{L-1}}\theta_L^k)\right)\cdot\nabla_{\theta_{L-1}}Z_{L-1}. \label{chain_theta}
\end{align}
Therefore, we can apply averaging on $\{\nabla_{Z_{L-1}}e^1,..,\nabla_{Z_{L-1}}e^K\}$, and calculate the gradient for $\theta_{L-1}$ once. The gradients for each $\theta_L^k$ must still be calculated separately with respect to $e^k$ (Algorithm \ref{algo_2}).
\section{Experiments}
\subsection{ACE for multiple models}
For the vanilla version we conducted an experiment over the MNIST dataset. The MNIST is a standard toy dataset, where the task is to classify the images into 10 digit classes. For the ensemble, 
\begin{wrapfigure}{r}{0.6\textwidth}
    \begin{minipage}{0.6\textwidth}
      \begin{table}[H]
  \caption{ACE for multiple models - MNIST dataset}
  \centering
  \begin{tabular}{lllll}
    \toprule
    &\multicolumn{2}{c}{Ensemble scores}  & \multicolumn{2}{c}{Averaged single NN score}   \\
    \cmidrule(r){1-5}
    $\lambda$     & Accuracy     & CE  & Accuracy     & CE \\
    \midrule
    \textbf{0}    & 0.9790 & 0.0669 & 0.9767 & 0.0810\\
    \textbf{0.05} & 0.9798 & 0.0663 & \textbf{0.9770} & 0.0809\\
    \textbf{0.1}  & 0.9799 & 0.0664 & 0.9768 & \textbf{0.0802}\\
    \textbf{0.3}  & 0.9797 & 0.0658 & 0.9767 & 0.0806\\
    \textbf{0.5}  & \textbf{0.9802} & \textbf{0.0649} & 0.9764 & 0.0842\\
    \textbf{0.7}  & 0.9800 & 0.0659 & 0.9760 & 0.0866\\
    \bottomrule
  \label{MNIST_table}
  \end{tabular}
\end{table}
    \end{minipage}
  \end{wrapfigure}
  we used 5 models of the same architecture. The architecture was DNN with a single hidden layer and ReLU activation. The results include both the accuracy and the CE of the predictions over the test set. We ran over multiple values of $\lambda\in [0,1]$, where for $\lambda=0$,  i.e. vanilla CE, we trained the models independently (different training batches). The results in Table \ref{MNIST_table} show that we succeeded in reducing the error of the ensemble and increasing its accuracy by applying ACE instead of the vanilla CE (i.e. $\lambda>0$). We also added the averaged accuracy and CE of a single DNN. An interesting thing to notice is that even though the result of a single DNN deteriorates when using the optimal $\lambda$, the ensemble result is superior. The reason for this is that we add a penalty for each DNN during the training phase that causes it to perform worse; however, the penalty is coordinated with the other DNNs so that the ensemble would perform better. The results were averaged over 5 experiments.
\subsection{Stacked Mixture Of Classifiers}
We conducted studies of the SMOC architecture over the CIFAR-10 dataset \cite{krizhevsky2009learning}. We used the architecture and code of ResNet 110 \cite{he2016deep} and stacked on top of it an ensemble of 10 fully-connected+Softmax layers instead of the single one that was used. This resulted in adding $5850$ parameters to a model with an original size of $1731002$, i.e. enlarging the model by $0.34\%$. The results are shown in Table \ref{CIFAR-10_table}. In the table we also show the results for a single classifier with a vanilla single Softmax layer (K=1). The results have been averaged over 5 experiments with different seeds. We notice that the optimal $\lambda$ reduces the accuracy error by $\sim 7\%$ compared to $K=1$ with almost no cost in the number of parameters and computational power. We also notice that the CE reduces significantly.
\begin{table}[!ht]
  \caption{Stacked Mixture Of Classifiers - CIFAR-10 dataset}
  \label{CIFAR-10_table}
  \centering
  \begin{tabular}{lllllllll}
    \toprule
    $K$  & $1$  & $10$       &  $10$           & $10$          & $10$          & $10$         & $10$         & $10$         \\ 
    $\lambda$        &        & $0$  &  $0.001$  & $0.01$  & $0.05$  & $0.1$  & $0.3$  & $0.5$  \\ 
    \midrule
      \textbf{error(\%)}   &  6.43 & 6.2          &   6.14            &  6.12           & \textbf{5.98}   &  6.09          &  6.13          &  6.31          \\
      {\textbf{CE}} & 0.3056& 0.3102       & 0.3041            & 0.3048          & 0.2968          & \textbf{0.2918}& 0.3137         & 0.4957         \\
    \bottomrule
  \end{tabular}
\end{table}
\section{Conclusion and future work}
In this paper we developed a novel framework for encouraging diversity explicitly between ensemble models in classification tasks. First, we introduced the idea of using an amended cost function for multiple classifiers based on NCL results. Later, we showed two usages - a vanilla one and the SMOC. We perform experiments to validate our analytical results for both of the architectures. For SMOC, we showed that by a small change and redundant addition of parameters we achieve superior results compared to the vanilla implementation. In future work, we would like to seek a way of using ACE with a non-uniform and, possibly, trainable weights \eqref{non_uniform_ACE}. Also, in the case of a large amount of labels, using SMOC results in a high amount of added parameters. We would like to research implementation solutions where this can be avoided.
\newpage
\bibliographystyle{unsrtnat}
\bibliography{main}
\end{document}